\pdfoutput=1

\documentclass[11pt]{article}

\usepackage[preprint]{acl}

\usepackage{times}
\usepackage{latexsym}

\usepackage[T1]{fontenc}

\usepackage[utf8]{inputenc}

\usepackage{microtype}

\usepackage{inconsolata}

\usepackage{graphicx}

\usepackage{amsmath}
\usepackage{multirow}
\usepackage{booktabs}
\usepackage{bbm}
\usepackage{subfigure}
\usepackage{amssymb}
\makeatletter
\renewcommand{\maketag@@@}[1]{\hbox{\m@th\normalsize\normalfont#1}}\makeatother

\title{\raisebox{-0.2cm}{\includegraphics[width=1cm]{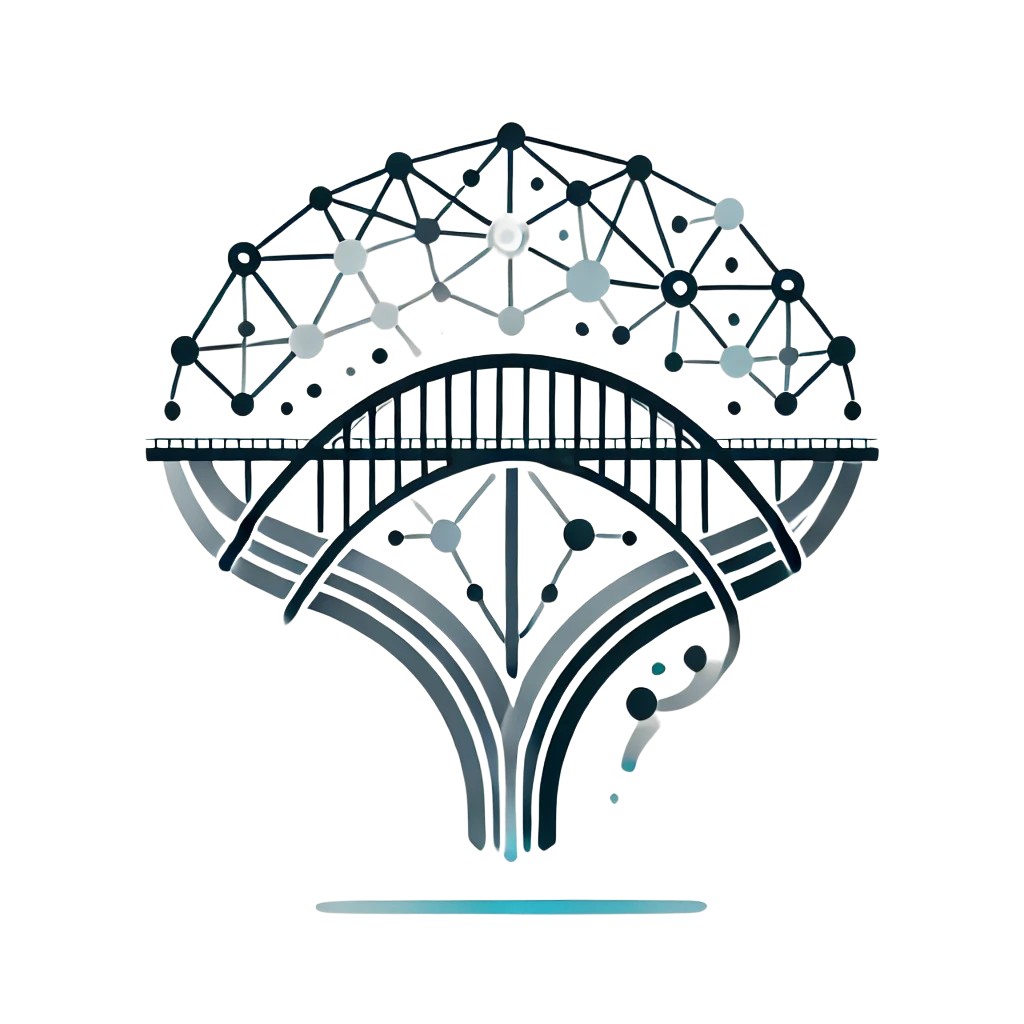}}
MindBridge: Scalable and Cross-Model Knowledge Editing via Memory-Augmented Modality}

\author{Shuaike Li\textsuperscript{1}, 
Kai Zhang\textsuperscript{1}, 
Qi Liu\textsuperscript{1}, 
Enhong Chen\textsuperscript{1}
\\ 
\textsuperscript{1}State Key Laboratory of Cognitive Intelligence, University of Science and Technology of China\\
\texttt{lishuaike767@gmail.com}\\
\texttt{\{kkzhang08, qiliuql, cheneh\}@ustc.edu.cn}}

\begin{document}
\maketitle

\begin{abstract}
Knowledge editing is a technique for efficiently and accurately updating the knowledge of large language models (LLMs) to alleviate obsolescence and correct errors. However, most existing methods overfit to specific models, causing edited knowledge to be discarded during each LLM update and requiring frequent re-editing, which is particularly burdensome in today's rapidly evolving open-source community. To address this issue, we propose the problem of cross-model knowledge editing and introduce \textbf{MindBridge}, a scalable solution inspired by the low coupling between modality processing and LLMs in multi-modal models. MindBridge introduces the novel concept of \textbf{memory modality}, which encodes edited knowledge as an independent modality. It first performs LLM-agnostic pre-training of the memory modality and then integrates it with various LLMs. Extensive experiments on multiple LLMs and popular knowledge editing datasets demonstrate that MindBridge achieves superior performance even in editing tens of thousands of knowledge entries and can flexibly adapt to different LLMs. Our code is available at \href{https://github.com/CrashBugger/MindBridge}{https://github.com/CrashBugger/MindBridge}.
\end{abstract}

\section{Introduction}
Large language models (LLMs) have revolutionized natural language processing, demonstrating remarkable abilities in understanding and generation \cite{achiam2023gpt, touvron2023llama, brown2020language}. These models leverage the knowledge acquired during pre-training to answer user queries. However, since the knowledge embedded in LLMs is stored in static parameters, their internal knowledge needs to be updated to keep pace with the ever-changing world and avoid obsolescence. Additionally, for personalized user information or domain-specific knowledge, customizing LLM output also requires updating the model's knowledge. Traditional methods, such as fine-tuning, continual learning, or retraining, are computationally expensive and may inevitably degrade the model's general capabilities\cite{kalajdzievski2024scaling,wang2023trace}. Fortunately, recent advancements in knowledge editing offer a promising solution, enabling efficient and precise modification of model knowledge at a low computational cost.

\begin{figure}[t]
    \centering 
    \vspace{0.25cm}
    \includegraphics[width=0.95\linewidth]{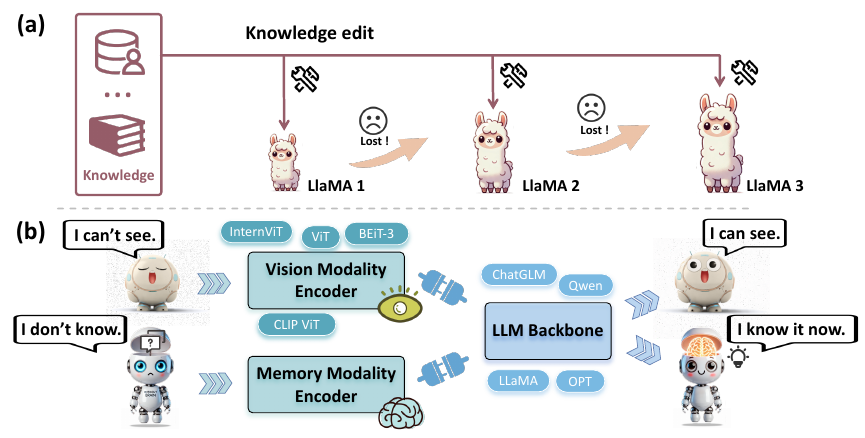}
        \caption{(a) The cross-model knowledge editing problem. Current knowledge editing methods discard previously edited knowledge after every LLM update (e.g., when the base model is updated alongside LLaMA), requiring frequent re-editing, which is labor-intensive. \emph{\textbf{This motivates us to explore whether edited knowledge can transcend individual models, i.e., achieve cross-model knowledge editing}}. (b) Analogizing memory modality to visual modality. Different visual modality encoders exhibit low coupling with various LLM backbones, and after efficient modality alignment, they enable LLMs to see. \emph{\textbf{Inspired by this, we propose \textit{memory modality}, which decouples knowledge from a single model, allowing knowledge editing through modality bridging with different LLMs.}}}
    \label{fig:intro}
    \vspace{-0.25cm}
\end{figure}

Existing knowledge editing methods generally fall into two categories \cite{yao2023editing}. The first one preserves LLM parameters, such as memory-based methods \cite{mitchell2022memory,madaan2022memory} or methods that add extra parameters to the model \cite{T-patcher, dong2022calibrating}. The second category modifies the parameters of the edited model, such as meta-learning-based methods \cite{mend,MALMEN} or locate-and-edit methods \cite{ROME, alphaedit}. While these methods have made significant progress, most still face the issue of overfitting to a single LLM. If the knowledge to be edited is domain- or user-specific, re-editing this knowledge becomes necessary whenever the LLM is updated, which is both tedious and time-consuming, especially given the fast pace of updates in the open-source community. This motivates us to explore whether the edited knowledge can be loosely coupled with the target model and is no longer limited to a single LLM. We call this problem {\emph{cross-model knowledge editing}}, as shown in Figure \ref{fig:intro}(a).

To address this issue, inspired by the work on multimodal large language models, we introduce the \textbf{memory modality}, a novel concept that encodes edited knowledge as a standalone modality to partially decouple a model's knowledge from the LLM itself. As illustrated in Figure~\ref{fig:intro}(b), the analogy to visual modality offers a clearer understanding of memory modality. The current mainstream paradigm\cite{llava} involves concatenating pre-trained visual modality encoders (e.g., ViT~\cite{VIT}, CLIP ViT~\cite{CLIP}) with various LLM backbones (e.g., LLaMA, Vicuna~\cite{vicuna}) to enable LLMs with visual perception. The low-coupling characteristic between visual modality processing and LLMs allows for independent updates of both components \cite{internvl, Vcoder}. This low-coupling property is precisely what is needed to solve the problem of cross-model knowledge editing. By encoding editable knowledge as a separate memory modality, pre-training a memory modality encoder to handle this knowledge, and subsequently concatenating it with multiple LLMs, we achieve efficient cross-model knowledge editing.

Based on the idea of memory modality, we propose \textbf{MindBridge}, a two-stage solution for cross-model knowledge editing. The first stage is \textit{memory modality pre-training}, where we introduce three training objectives: memory injection, memory association, and memory existence. These objectives are designed to train the memory modality encoder so that it can acquire relevant memories, perform memory association, and determine whether certain memories exist. The second stage is \textit{memory modality bridging}, where we fine-tune a simple projector to achieve efficient cross-modal alignment, enabling the output of the memory modality to be understood and utilized by LLMs. This approach allows for efficient large-scale knowledge updates, with minimal modifications required when the LLMs are updated.

To test the effectiveness of MindBridge, we conduct extensive experiments on widely used knowledge editing datasets, ZsRE and Counterfact, using multiple LLMs, including GPT-XL, GPT-J, and LLaMA3. Leveraging the low coupling characteristic between the memory modality and LLMs, MindBridge achieves efficient and scalable knowledge editing across different models, delivering excellent editing performance even in scenarios involving tens of thousands of knowledge edits. Our contributions can be summarized as follows:

\begin{itemize}
\vspace{-0.2cm}
    \item We introduce the novel problem of \emph{cross-model knowledge editing}, addressing the critical challenge of discarding previously edited knowledge and repeatedly re-editing caused by the rapid iteration of LLMs.
    \vspace{-0.2cm}
    \item We propose MindBridge, a scalable and effective solution that leverages the innovative concept of a memory modality to achieve cross-model knowledge editing.
    \vspace{-0.2cm}
    \item Extensive experiments on knowledge editing datasets and multiple LLMs validate the effectiveness and scalability of MindBridge for cross-model knowledge editing, even when handling tens of thousands of edits.
\end{itemize}

\section{Related Work}

\subsection{Knowledge Editing}

Knowledge editing, aimed at inserting new knowledge or modifying existing knowledge in LLMs to alter their behavior, falls into two main paradigms \cite{yao2023editing}.  \textit{Parameter-modifying editing} directly adjusts model parameters. This includes locate-then-edit strategies like ROME \cite{ROME}, MEMIT \cite{memit}, and Alphaedit \cite{alphaedit}, which pinpoint relevant knowledge before fine-tuning parameters.  Meta-learning approaches, such as KE \cite{KE}, MEND \cite{mend}, and MALMEN \cite{MALMEN}, also belong to this paradigm, training hypernetworks to generate parameter updates.

In contrast, \textit{parameter-preserving editing} maintains original model weights and uses external components for knowledge storage. T-Patcher \cite{T-patcher} adds neurons to the last feed-forward layer to adjust output. SERAC \cite{SERAC} trains a counterfactual model using a classifier to determine response relevance.  Database-retrieval methods, including GRACE \cite{grace} (hidden vector database) and MELO (LoRA block database), modify LLM computation based on retrieved knowledge.  MemPrompt \cite{memgpt} and IKE \cite{IKE} utilize retrieved demonstrations as prompts, leveraging in-context learning.

While effective for targeted edits, existing knowledge editing methods are often overfit to specific LLMs and struggle with cross-model, large-scale knowledge editing.  After each iteration of the LLM, the previously edited knowledge is lost, leading to the need for frequent re-editing.

\subsection{MultiModal Large Language Models}

Recent advancements in multimodal large language models (MM-LMMs) have enabled LLMs to process diverse modality inputs like images, video, and audio \cite{blip2,videochat,qwen-audio}. Due to the high cost of training MM-LLMs from scratch, a more efficient strategy involves integrating pre-trained unimodal foundation models with the LLM backbone. This approach typically follows a two-stage pipeline: Multimodal Pre-Training, which aligns modality encoder features with the LLM, and Multimodal Instruction-Tuning, which ensures instruction following and zero-shot generalization \cite{zhang2024mm,llava}.

This decoupling of the LLM backbone from modality encoders enables the reuse or iterative updates of modality encoders without affecting the LLM \cite{internvl,dinov2}, offering insights for cross-model knowledge editing. Beyond traditional modalities, we propose the novel concept of memory modality to achieve similar decoupling, partially separating model knowledge and reasoning. This enables efficient knowledge editing, allowing for the retention or independent updating of previously edited knowledge even when the LLM backbone is updated.

\section{Problem Formulation}
The goal of knowledge editing is to modify the knowledge stored in a model. In this paper, we focus specifically on editing memories composed of factual knowledge. More concretely, for a triplet $(s, r, o)$ consisting of a subject $s$, relation $r$, and object $o$ (e.g., $s$ = United States, $r$ = President, $o$ = Biden), we aim to insert a new triplet $(s, r, o^*)$ (e.g., $s$ = United States, $r$ = President, $o^*$ = Trump) into the LLM to replace the previous knowledge, i.e., $(s, r, o) \to (s, r, o^*)$, where these two triplets share the same subject and relation. Specifically, $o^*$ can also represent knowledge that does not originally exist in the LLM, i.e., $(s, r, \emptyset) \to (s, r, o^*)$.

Given a set of knowledge to be edited $D_{\text{edit}} = \{(s_i, r_i, o_i^*) \mid i = 1, 2, \ldots, n\}$, a knowledge editing operation $KE$, and a model to be edited $F$, the goal of knowledge editing is to generate a new model $F^*$ through the knowledge editing operation $KE$. This can be formulated as follows:

\begin{small}
    \begin{equation} \label{equation edit}
\begin{aligned}
F^* &= KE(F, D_{edit}), \\
\text{s.t.} \; F^*(s, r) &= 
\begin{cases}
o^*, & \text{if } (s, r,o^*) \in \mathcal{I}(D_{edit}), \\
F(s, r), & \text{if } (s, r,o^*) \in \mathcal{O}(D_{edit}).
\end{cases}
\end{aligned}
\end{equation}
\end{small}

Here, $\mathcal{I}(D_{edit})$ denotes the knowledge set that requires editing and its neighborhood, such as paraphrasing and rewriting, with $D_{edit} \subseteq \mathcal{I}(D_{edit})$. Meanwhile, $\mathcal{O}(D_{edit})$ denotes the set of knowledge items unrelated to the edited knowledge. For simplicity, we will refer to these sets as $\mathcal{I}$ and $\mathcal{O}$ moving forward. In some studies, $\mathcal{I}$ is also referred to as in-scope examples, while $\mathcal{O}$ is termed out-of-scope examples \cite{SERAC}.

Equation \ref{equation edit} specifies that the edited model $F^*$ should correctly predict the edited knowledge and its neighborhood. For inputs unrelated to the edited knowledge, $F^*$ should maintain consistent predictions with the original model $F$. This ensures that knowledge editing updates target knowledge while minimizing interference with unrelated knowledge.

\section{Method}

\begin{figure*}[htp]
    \centering
        \includegraphics[width=1\linewidth]{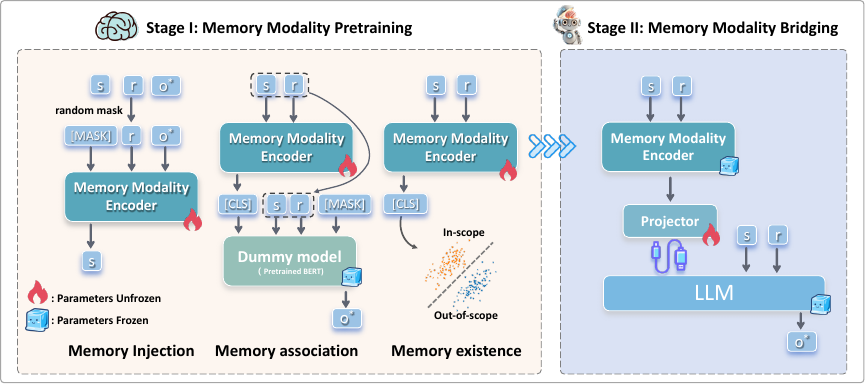}
            \caption{Overview of MindBridge. Given a massive collection of fact knowledge subject-relation-object triplets $(s, r, o^*)$ intended for editing, we first perform \emph{\textbf{stage I: Memory Modality Pre-training}}. This phase utilizes three training objectives -- memory injection, memory association, and memory existence -- to develop a memory modality encoder capable of retaining relevant memories, making associations and determining whether specific memories exist. In \emph{\textbf{stage II: Memory Modality Bridging}}, we then train a projector to bridge the memory modality encoder with LLMs, allowing these models to obtain and effectively leverage the relevant memories.}
    \label{fig:overview}
\end{figure*}

As illustrated in Fig. \ref{fig:overview}, MindBridge is meticulously structured into two primary stages.  The first stage, \textbf{Memory Modality Pre-training} (detailed in Section \ref{Memory Modality Pre-training}), is specifically designed to obtain the memory modality encoder, $E_{m}$. The second stage, \textbf{Memory Modality Bridging} (detailed in Section \ref{Memory Modality Bridging}), is dedicated to training a projector to construct the memory-to-language bridging module, $P_{m}$.  Ultimately, the edited model, $F^*$, can be mathematically represented by Eq. \ref{equation edited model}.  In MindBridge, knowledge is encoded into memory modality features, denoted as $x_{memory}$, through the synergistic utilization of the memory modality encoder $E_m$ and the bridging module $P_m$. These features are subsequently concatenated with the textual input $(s,r)$ as soft prompts and provided to the LLM, $F$, for prediction.

\begin{equation}\label{equation edited model}
\begin{aligned}
    x_{memory} &= P_m(E_m(s,r)), \\
    F^*(s,r) &= F(x_{memory} \oplus (s,r)).
\end{aligned}
\end{equation}

 To fully appreciate the design of MindBridge, we first review the general architecture of current mainstream multimodal large language models (MM-LLMs) \cite{llava,qwen_vl}. These models typically consist of three shared design elements: a modality encoder $E$, a modality-to-language alignment module $P$, and a frozen LLM $F$. The final constructed MM-LLMs resemble the edited model presented in Eq.\ref{equation edited model}. During inference, MM-LLMs rely on the modality encoder $E$ and the bridging module $P$ to extract modality features and transform them into a text feature space understandable by LLMs. Drawing inspiration from this architecture's successful design, MindBridge adopts a similar architecture to implement the training and bridging of memory modalities. Consequently, the missing components in MindBridge are the modality encoder and the modality-to-language alignment module, which are addressed in its two stages respectively.

\subsection{Memory Modality Pre-training}\label{Memory Modality Pre-training}

In multimodal large language models, modality encoders are typically pre-trained in advance to extract features from modality-specific inputs, such as the commonly used vision modality encoders ViT \cite{VIT} and CLIP-ViT \cite{CLIP}.  Analogously, for the memory modality, pre-training a dedicated encoder, $E_m$, is indispensable. From the perspective of the knowledge editing workflow, this encoder should be designed to achieve the following three key objectives:

\begin{itemize}
    \item \emph{{Objective 1}}: Possess relevant memories for target knowledge, i.e., store tuples of $(s, r, o^*)\in D_{edit}$.
    \vspace{-0.1cm}
    \item \emph{Objective 2}: Extract relevant memories based on provided context, i.e., retrieve memories associated with $(o^*)$ from the given $(s, r)$, where $(s, r, o^*) \in D_{edit}$.
    \vspace{-0.1cm}
    \item \emph{Objective 3}: Distinguish whether relevant memories exist, i.e., to differentiate between the set $\mathcal{I}$ and the set $\mathcal{O}$.
\end{itemize}

We initialize $E_{m}$ using a pre-trained BERT \cite{bert} and use the [CLS] hidden state of its output as the extracted memory modality features, i.e., $E_m(s, r) = \text{BERT}_{\text{[CLS]}}(s, r)$. To achieve Objective 1, we adopt the Masked Language Model (MLM) training objective to inject the knowledge $(s, r, o^*)\in D_{edit}$ into $E_{m}$. Specifically, we randomly mask one element of the $(s, r, o^*)$ triplet by replacing it with the [MASK] token and then have $E_{m}$ reconstruct the masked element. We refer to this loss function as the memory-injection loss $L_{inject}$, which can be expressed as follows:

\begin{multline*}
    L_{\mathrm{inject}}=-\mathbb{E}_{(s,r,o^*)\sim D_{\mathrm{edit}}} \\ \big[\sum_{x_i \in\{s,r,o^*\}}  \log  \mathbbm{P}_{E_m}(x_i \mid[\mathrm{MASK}],(s,r,o^*)\setminus {x_i})\big].
\end{multline*}

To achieve Objective 2, we need to enhance the memory modality features extracted from the [CLS] token so that they contain representations related to $o^*$ based on $(s, r)$. This process mirrors human associative memory, where observing a segment of text triggers recall of associated information. To adapt the memory modality encoder for this task, we feed the [CLS] representation output by $E_m(s, r)$ along with $(s, r, [\text{MASK}])$ into another dummy model $M$. The model $M$ must predict $o^*$ based on the associations made by the memory modality given $(s, r)$. In this case, $M$ is also a BERT model, sharing the same pre-trained initialization parameters with $E_m$, but its parameters remain frozen throughout the training process. We refer to this loss function as the memory-association loss $L_{\text{associate}}$, which can be formulated as follows:
\begin{multline*}
    L_{\mathrm{associate}} = - \mathbb{E}_{(s,r,o^*) \sim D_{\mathrm{edit}}} \\ \left[\log \mathbbm{P}_{M}(o^* \mid E_m(s,r)\oplus (s,r,[\mathrm{MASK}])) \right].
\end{multline*}

To achieve Objective 3, which involves determining whether relevant memories exist, we need to ensure that $E_m(s, r)$ produces distinct representations for the sets $\mathcal{I}$ and $\mathcal{O}$. We adopt a simple binary classification task to accomplish this goal. Specifically, we feed $E_m(s, r)$, where $(s, r, o^*) \in \mathcal{I} \cup \mathcal{O}$, into a classification head $H$ composed of two linear layers to classify whether the input belongs to $\mathcal{I}$ or $\mathcal{O}$. This approach ensures that the internal representations of $\mathcal{I}$ and $\mathcal{O}$ are clearly differentiated and reside in distinct vector spaces. We refer to this loss function as the memory-existence loss $L_{exist}$, which can be expressed as follows:
\begin{align*}
\hat{y} &= H(E_m(s, r)),\\
L_{exist} &= -\mathbb{E}_{(s, r,o^*) \sim \mathcal{I} \cup \mathcal{O}} \Big[ y \log(\hat{y}) \\
&\quad + (1 - y) \log(1 - \hat{y}) \Big],
\end{align*}
where $y$ is the true label indicating membership in $\mathcal{I}$ ($y=1$) or $\mathcal{O}$ ($y=0$).

Finally, the overall loss function for memory modality pre-training is shown in Equation \ref{loss pretrain}, where $\lambda_1$, $\lambda_2$, and $\lambda_3$ are coefficients for different loss functions, with default values set to 1. Notably, the memory modality pre-training phase is entirely independent of the LLMs $F$ that are to be edited. Therefore, it can be trained independently to embed a substantial amount of knowledge memory. As LLMs are updated or replaced over time, $E_m$ can be reused across various models, facilitating efficient cross-model editing.
\begin{equation}
\label{loss pretrain}
    L_{ \text{pre-training}} = \lambda_1 L_{\text{inject}} +  \lambda_{2} L_{\text{associate}} \\ +  \lambda_{3} L_{\text{exist}}.
\end{equation}

\subsection{Memory Modality Bridging}\label{Memory Modality Bridging}
After memory modality pre-training, we have obtained a memory modality encoder $E_m$, which has stored the relevant memories of $(s, r, o^*)$ and is capable of extracting related memories and determining their existence based on context. Revisiting Equation \ref{equation edited model}, we observe that we still need a memory-to-language module $P_m$ to bridge the memory modality to LLMs. This module allows LLMs to understand and interpret the memories.

In mainstream MM-LLM works \cite{blip2,llava,P-former}, to obtain the modality-to-language alignment modules $P$, a modality-conditioned text generation loss is typically used as the training objective. Drawing inspiration from this, we also adopt this training objective, enabling the LLM $F$ to correctly predict $o^*$ based on the output of the memory modality and the prompt (composed of $s$ and $r$), as specifically shown in Equation \ref{equa l1}. Here, we use a simple two-layer fully connected network as $P_m$, while keeping the parameters of the LLM $F$ and the memory modality encoder $E_m$ frozen.
\begin{multline}
    L_1 = - \mathbb{E}_{(s, r, o^*) \sim D_{\text{edit}}} \\ \left[\log \mathbbm{P}_{F}(o^* \mid P_m( E_m(s,r)) \oplus(s,r))\right].
\label{equa l1}
\end{multline}

However, training solely with this objective only enables LLMs to comprehend knowledge present in memory. For knowledge not contained in memories, the output from the memory modality may lead to unpredictable behavior. Ideally, for knowledge not present in memories, i.e., $(s, r, o^*) \in \mathcal{O}$, we expect the LLMs to ignore the memory modality's output and maintain their original predictions. To achieve this, we minimize the discrepancy between pre- and post-edit model predictions by reducing the Kullback-Leibler (KL) divergence of their prediction distributions\cite{kl_div}, as illustrated in Equation \ref{equ kl}. Benefiting from the pre-trained memory modality’s ability to differentiate representations of in-scope ($\mathcal{I}$) and out-of-scope ($\mathcal{O}$) knowledge, LLMs can more readily discern whether they possess the relevant memory.
\begin{multline}
    L_2 = \mathbb{E}_{(s, r,o^*) \sim \mathcal{O}} \left[ \text{KL}\left( \mathbbm{P}_{F}(\cdot \mid (s,r)) \right. \right. \\ \left. \left. \parallel \mathbbm{P}_{F^*}( \cdot \mid P_m( E_m(s,r)) \oplus(s,r)) \right) \right.].
\label{equ kl}
\end{multline}

In summary, we combine the two training objectives of the bridging phase into a single loss function, as shown in Equation \ref{equ all}, where $\lambda_{\text{kl}}$ is the coefficient, defaulting to 1. Although the memory modality bridging phase is not LLM backbone agnostic, it only requires fine-tuning $P_m$. Given that the parameters involved in this fine-tuning are negligible compared to the total number of parameters, it enables an efficient and rapid implementation of memory modality bridging.

\begin{equation}
     L_{\text{bridge}} = L_1 + \lambda_{kl}L_2.
\label{equ all}
\end{equation}

\section{Experiments}

\begin{table*}[t]\label{Baseline table}
\centering
\small
\begin{tabular}{cc|cccc|cccc}
\toprule
\multirow{2}{*}{\textbf{Method}} & 
\multirow{2}{*}{\textbf{Model}} & 
\multicolumn{4}{c|}{\textbf{Counterfact}} & 
\multicolumn{4}{c}{\textbf{ZsRE}} \\
\cmidrule(lr){3-6} \cmidrule(lr){7-10} & & 
\textbf{Rel.$\uparrow$} & \textbf{Gen.$\uparrow$} & \textbf{Loc.$\uparrow$} & \textbf{Avg.$\uparrow$} &  
\textbf{Rel.$\uparrow$} & \textbf{Gen.$\uparrow$} & \textbf{Loc.$\uparrow$} & \textbf{Avg.$\uparrow$} \\

\midrule
 Pre-edited &  & 0.28 & 0.42 &  \textbackslash  &\textbackslash & 22.44 & 21.56 &\textbackslash & \textbackslash \\
\midrule 			
			
  FT-L  &   \multirow{7}{*}{\rotatebox{90}{GPT-XL}}  & 0  & 0 &  22.84 & 7.61 &  15.60 & 15.83 & 28.21 & 19.88 \\
  WISE  &     & 1.79 & 1.99 & 50.38 & 18.06 & 24.88 & 24.84 & \underline{99.98} & 49.90\\
 AlphaEdit & & 66.00 & 30.90 & 42.98 & 46.62 &  51.82 & 42.97 & 53.72 &   49.50 \\
  EMMET   &   & 85.51  &  \underline{46.65} & 62.17 & \underline{64.78} & 70.37 & \underline{60.34} & 78.37 &  \underline{69.69} \\
 GRACE   &    & \textbf{100} & 0.39 & \textbf{ 68.75} & 56.38 &  \textbf{100} & 3.17 & \textbf{100}  & 67.72 \\
  r-ROME   &    & 0 & 0 & 0 & 0 & 0 &  0&  0& 0\\
  MEMIT  &    & 45.36 &  39.70& \underline{66.25} &  44.08 & 47.20 & 24.70 & 60.34 & 50.44 \\
  			
  MindBridge  &     &  \underline{95.60} & \textbf{81.10} &  38.18 & \textbf{71.62} & \underline{78.14} & \textbf{68.17}  & 85.67 & \textbf{77.33} \\
\midrule

\midrule
  Pre-edited &    & 0.09 & 0.29 &  \textbackslash & \textbackslash & 23.01  & 22.25 & \textbackslash & \textbackslash \\
\midrule
  FT-L  &   \multirow{7}{*}{\rotatebox{90}{GPT-J}}  & 10.58 & 7.39 &  1.03 &  6.34 & 15.73 & 14.04  & 9.47  & 13.08 \\
  WISE  &     & 18.38  & 12.08 & 3.85 &  11.44 & 36.88 & 34.53  & \underline{99.51}  & 56.97 \\
 AlphaEdit & &  92.00 & 46.45 & 57.40 & 65.28  & 76.95 & 52.37 &62.02  & 63.78 \\
  EMMET   &   & 85.51 & 46.65 & 62.17 & 64.78 & 70.37 & 60.34 & 78.37 & 69.69 \\
 GRACE   &    & \textbf{100} & 0.29 & \textbf{99.08} & 66.46 & \textbf{100} & 3.13 & \textbf{100} & 67.71\\
  r-ROME   &    &  0& 0 & 0 & 0 &  0.09 & 0.08  &  0 & 0.05\\		
  MEMIT  &    & \underline{95.80}  & \underline{57.64} & 61.87  & \underline{71.77} & 86.17 & \underline{69.70} & 75.96 & \underline{77.28} \\

  MindBridge  &     &94.60  & \textbf{82.60} & \underline{93.56} & \textbf{90.25} & \underline{99.06} & \textbf{83.71} & 95.77  & \textbf{92.85} \\
\midrule

\midrule
Pre-edited &  & 0.7  & 1.3 &  \textbackslash & \textbackslash & 27.70  & 27.08 & \textbackslash & \textbackslash \\
\midrule

  FT-L  &   \multirow{7}{*}{\rotatebox{90}{LLaMA3}}   & 25.67 & 9.24 & 0.23 & 11.71 & 5.99 & 3.92  & 0.64 & 3.52 \\
  WISE  &     & 16.08 & 10.68 &3.87  & 10.21 &29.18  & 29.18 & \underline{99.40} & 52.07 \\
 AlphaEdit & & 91.70 & \underline{51.54} & 55.58 & 66.28 & 78.64 & \underline{ 62.06} &73.57  & \underline{71.43}  \\
  EMMET   &   & 60.53 & 31.81 & 32.28 & 41.54 & 62.94 & 59.64 & 31.95 & 51.51 \\
 GRACE   &    & \textbf{100}  & 5.33 & \textbf{100} & \underline{68.44} & \textbf{100} & 5.33 & \textbf{100} & 68.44 \\
  r-ROME   &    & 0.09 & 0 & 0.15 & 0.08 & 0.70 & 0.50 & 1.01 &  0.74 \\
  MEMIT  &    & 84.71 & 41.20 & 64.49 & 63.47 & 66.16 &  59.92 & 80.51 & 68.86 \\
	
  MindBridge  &     & \underline{93.85}  & \textbf{83.35}  & \underline{92.14} & \textbf{89.78} & \underline{99.03}  &\textbf{85.50 } & 92.65 & \textbf{92.39}\\
\bottomrule
\end{tabular}
\caption{Comparison of MindBridge with existing methods after editing 10,000 facts. The best results are shown in bold, and the second best results are underlined.}
\label{tab baseline com}
\end{table*}

\subsection{Experimental Setup}
We first provide a brief overview of the datasets, models, metrics, and baseline methods used in our experiments. For more detailed information on the experimental setup, please refer to Appendix \ref{sec:Implementation Details}.

\textbf{Models and Datasets. } We conducted experiments on three LLMs of varying sizes: GPT2-XL (1.5B) \cite{GPT2-XL}, GPT-J (6B) \cite{gpt-j}, and LLaMA3 (8B) \cite{llama3}. For the benchmark, we evaluated MindBridge on two commonly used knowledge editing datasets: the ZsRE dataset \cite{zsre-dataset} and the Counterfact dataset \cite{ROME}. For MindBridge, we designated the knowledge in the dataset that required editing as $D_{edit}$, and a portion of the knowledge that did not require editing as $\mathcal{O}$. Since the quantity of $\mathcal{I}$ in the dataset is small and typically used only for testing, we directly substitute  $D_{edit}$ for  $\mathcal{I}$ during training.

\textbf{Baseline Methods.} We compare MindBridge with seven baseline methods, categorized into two groups: parameter-modifying knowledge editing approaches—Fine-Tuning (FT-L) \cite{ROME}, r-ROME \cite{r-rome}, MEMIT \cite{memit}, EMMET\cite{emmet} and Alphaedit \cite{alphaedit}; and parameter-preserving knowledge editing methods—GRACE \cite{grace} and WISE \cite{wise}. For these baseline methods, we utilize EasyEdit \cite{easyedit} for replication and testing, applying the default parameter settings.

\textbf{Metrics. } Following previous work (\cite{alphaedit,easyedit}), we adopt three metrics to evaluate the performance of the edited model: \textbf{reliability} (edit success rate), \textbf{generalization} (paraphrase success rate), and \textbf{locality} (neighborhood success rate). These are abbreviated as Rel., Gen., and Loc., respectively. We further compute the average of these three metrics, denoted as Avg., to represent the overall editing performance.

\subsection{Main Results}\label{sec:main results}

Table \ref{tab baseline com} shows the performance comparison between MindBridge and existing knowledge editing methods after editing 10,000 facts. We can observe that, compared to baselines, MindBridge demonstrates superior performance across multiple LLMs, different datasets, and almost all metrics. For example, on LLaMA3 and GPT-J, MindBridge outperforms the best baseline by more than 20\% and 15\% respectively on the Avg. metric, which measures the comprehensive editing performance. On the Gen. metric, MindBridge significantly outperforms other editing methods in all experiments.  Furthermore, it is particularly important to note that MindBridge is the only editing method among these that can achieve cross-model editing. When the LLM is updated, MindBridge can retain a large amount of previously edited knowledge and quickly bridge the memory modality encoder to the new LLM, enabling rapid domain knowledge adaptation while maintaining excellent editing performance.

\subsection{Further Analysis}
\label{sec:Further Analysis}

\begin{table*}[t]\label{ablation table}
\centering
\small
\begin{tabular}{c|cccc|cccc}
\toprule
\multirow{2}{*}{\textbf{Traing objectives}} &  
\multicolumn{4}{c|}{\textbf{Counterfact}} & 
\multicolumn{4}{c}{\textbf{ZsRE}} \\
\cmidrule(lr){2-5} \cmidrule(lr){6-9} & \textbf{Rel.$\uparrow$} & \textbf{Gen.$\uparrow$} & \textbf{Loc.$\uparrow$} & \textbf{Avg.$\uparrow$} &  
\textbf{Rel.$\uparrow$} & \textbf{Gen.$\uparrow$} & \textbf{Loc.$\uparrow$} & \textbf{Avg.$\uparrow$} \\
\midrule 			
Pre-edited &0.2 & 0.5& \textbackslash & \textbackslash & 24.20 &22.93 & \textbackslash &\textbackslash \\
\midrule  		
$L_{inject}$ & 71.30 & 34.20& 41.16 & 48.88 & 94.89 & 75.28 &74.23 & 81.47\\
$L_{inject}+L_{associate}$& 83.70 &  63.30 & 73.12 & 73.37 & 97.75 & 84.01 & 85.18& 88.98 \\
$L_{inject}+L_{exist}$&75.40& 50.70 & 88.40 & 71.50 & 95.86& 81.45 & 93.28 & 90.20 \\
$L_{inject}+L_{associate}+L_{exist}$& \textbf{94.40} &  \textbf{80.50} & \textbf{89.26} & \textbf{88.05} & \textbf{99.40} & \textbf{84.87} & \textbf{96.15}& \textbf{93.47} \\
\midrule
\end{tabular}
\caption{Ablation study of the three training objectives in MindBridge's memory modality pre-training. Edited LLM: GPT-J; 10,000 facts. Best results are highlighted in bold.}
\vspace{-0.1cm}
\label{tab ablation}
\end{table*}

\textbf{Ablation Study for Memory Modality Pre-training.} In the memory modality pre-training stage, we designed three training objectives for the modality encoder, each corresponding to its intended function. To validate the effectiveness of these three training objectives, we conducted an ablation study to evaluate the impact of each objective. We performed knowledge editing on 10,000 factual statements using GPT-J. The results, as shown in Table \ref{tab ablation}, demonstrate that employing only the $L_{inject}$ objective already yields promising editing performance (with Rel. metric reaching 71.30\% on the Counterfact dataset and 94.89\% on the ZsRE dataset). However, the generalization and locality of the edits are limited. Upon incorporating the $L_{associate}$ training objective in addition to $L_{inject}$, the generalization capability of editing is enhanced (with the Gen. metric increasing by 29.1\%$\uparrow$ on the Counterfact dataset and 8.73\%$\uparrow$ on the ZsRE dataset).  Furthermore, integrating the $L_{exist}$ training objective alongside $L_{inject}$ improves the locality of editing (with the Loc. metric increasing by 47.24\%$\uparrow$ on the Counterfact dataset and 19.05\%$\uparrow$ on the ZsRE dataset). These findings validate the effectiveness of each of the three training objectives. Ultimately, by combining all three training objectives, MindBridge achieves the best editing performance.

\begin{figure}
    \centering
    \includegraphics[width=1\linewidth]{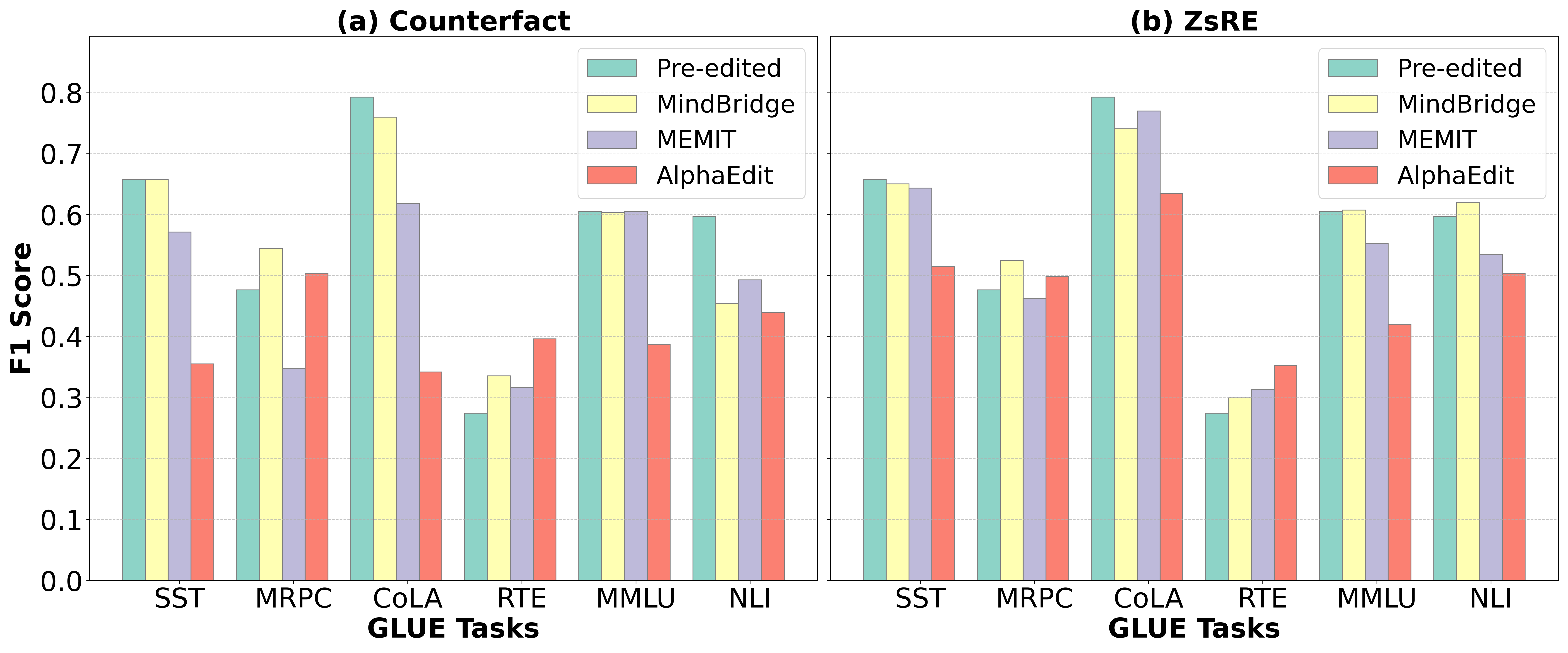}
    \caption{F1-score of LLaMA3 (8B) on the GLUE benchmark after editing 10,000 facts using MindBridge, AlphaEdit, and MEMIT. The evaluation includes six tasks: SST, MRPC, CoLA, RTE, MMLU, and NLI.}
    \label{fig:glue}
\end{figure}

\textbf{Impact on General Ability.} Existing knowledge editing methods may more or less affect the general capabilities of models \cite{gu2024model,gupta2024model}. To test the impact of MindBridge on the general capabilities of edited models, we evaluated the edited LLaMA3 (8B) on the General Language Understanding Evaluation (GLUE) benchmark \cite{glue_benchmark} and compared it with AlphaEdit and MEMIT, which exhibit good performance in comprehensive editing effectiveness. As shown in Figure \ref{fig:glue}, MindBridge has the smallest change in F1-score compared to Pre-edited on six tasks, indicating that MindBridge can maintain the general capabilities of the model even in the face of a large amount of knowledge editing.

\begin{figure}[t]
\centering
\vspace{-3mm}
\subfigure[Counterfact]{ \includegraphics[width=0.22 \textwidth]{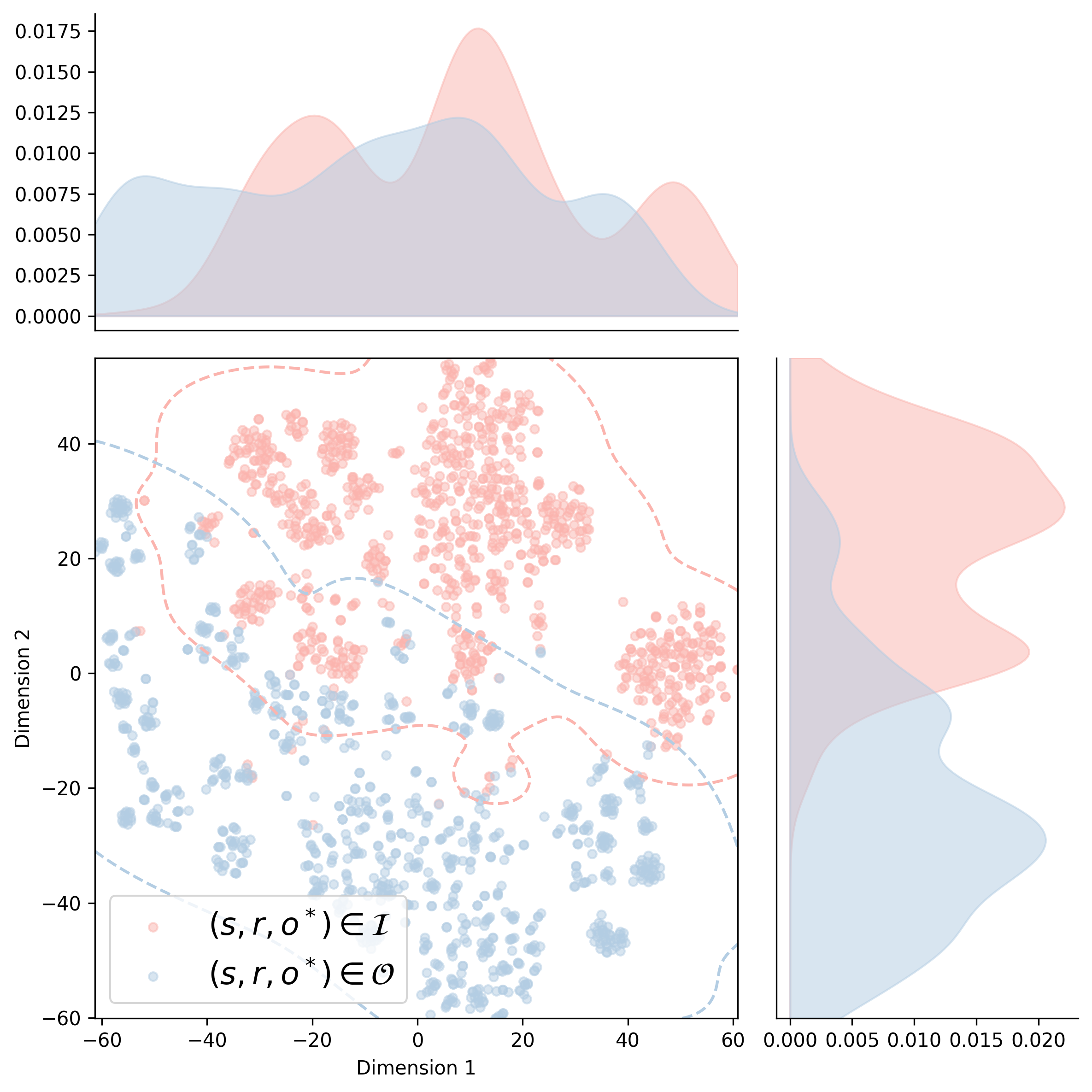}
}
\hspace{0.00001\textwidth}\subfigure[ZsRE]{ \includegraphics[width=0.22 \textwidth]{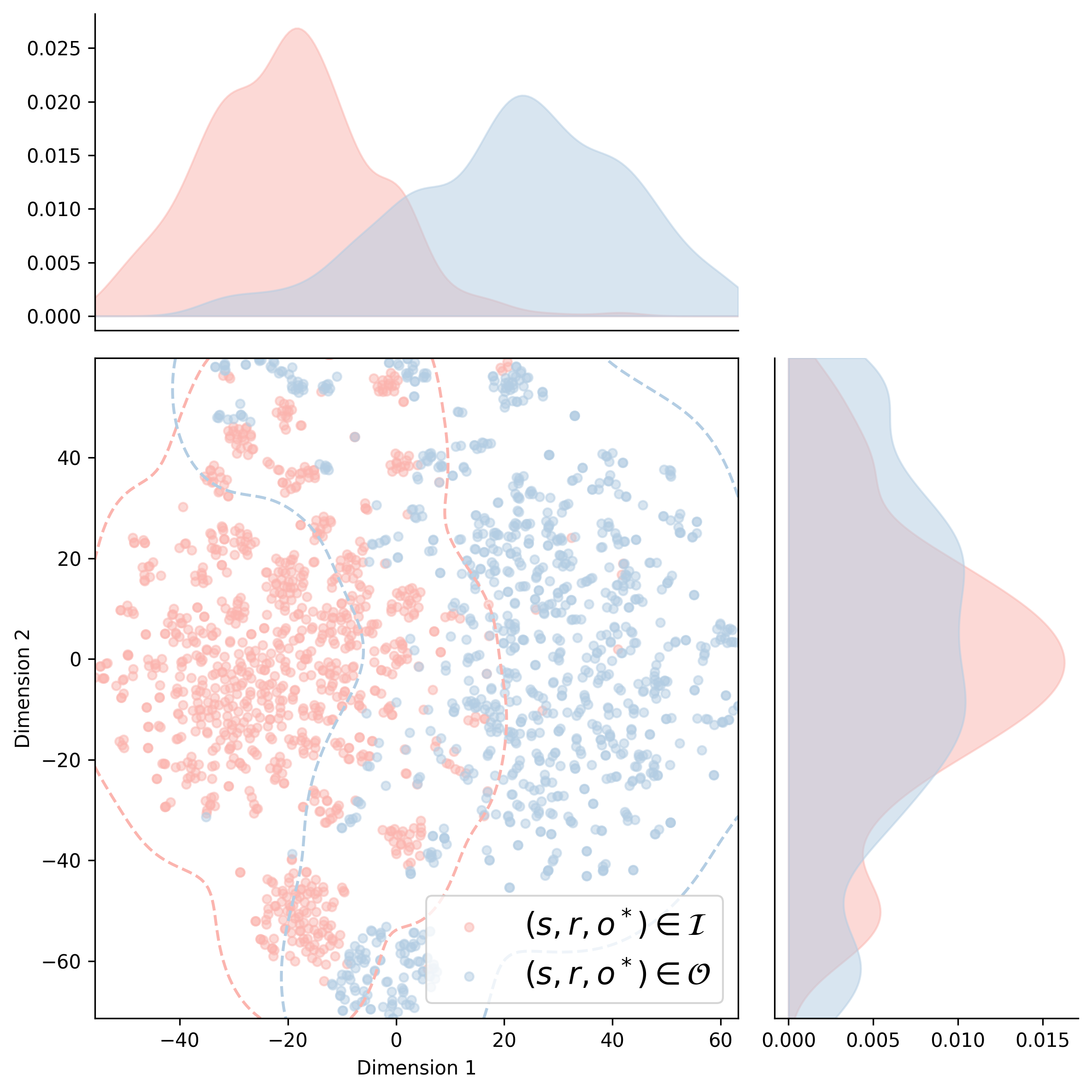}
}

\caption{Visualization of the dimensionality-reduced distributions of representations extracted by the memory modality encoder for $\mathcal{I}$ and $\mathcal{O}$. }
\label{fig:tsne}
\end{figure}

\textbf{Visualization of In-scope and Out-of-scope Representations.} To determine whether the memory modality can truly distinguish the presence of relevant memories, we randomly selected 1,000 examples from the in-scope and out-of-scope data of two datasets. Figure \ref{fig:tsne} shows the visualization of the representations extracted by the memory modality encoder, reduced to two dimensions using t-SNE. It can be observed that the data from $\mathcal{I}$ and $\mathcal{O}$ are separated in space, indicating that the memory modality encoder is capable of distinguishing whether relevant factual memories exist, thereby ensuring the locality of the edited LLM.

\begin{figure}[t]
\centering
\subfigure[GPT-J]{ \includegraphics[width=0.235 \textwidth]{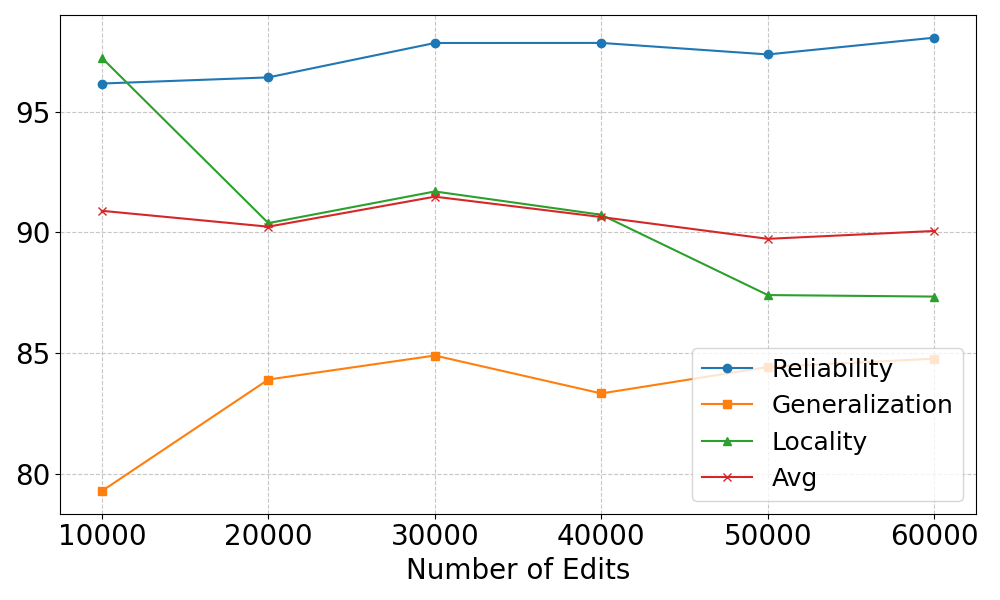}
}
\hspace{-0.02\textwidth}\subfigure[LLaMA3]{ \includegraphics[width=0.235 \textwidth]{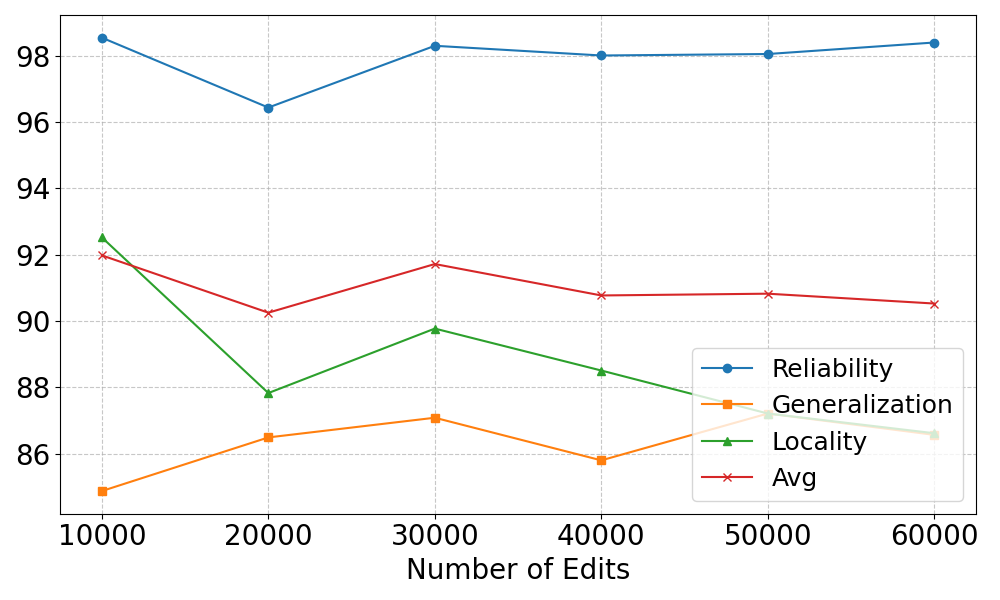}
}
\caption{Scaling MindBridge to 60,000 edits on ZsRE.}
\vspace{-0.25cm}
\label{fig:scalable}
\end{figure}

\textbf{Scale Up to 60,000 Edits.} We gradually extended MindBridge from 10,000 edits to 60,000 edits on the ZsRE dataset for GPT-J and LLaMA3, and tested their editing performance. As shown in Figure \ref{fig:scalable}, despite the significant increase in the number of edits, MindBridge consistently achieves stable and superior editing performance. The Avg. metric remains largely unchanged, with only a slight decrease observed in the Locality metric.

\begin{figure}[t]
\centering
\vspace{-3mm}
\subfigure[Counterfact]{ \includegraphics[width=0.22 \textwidth]{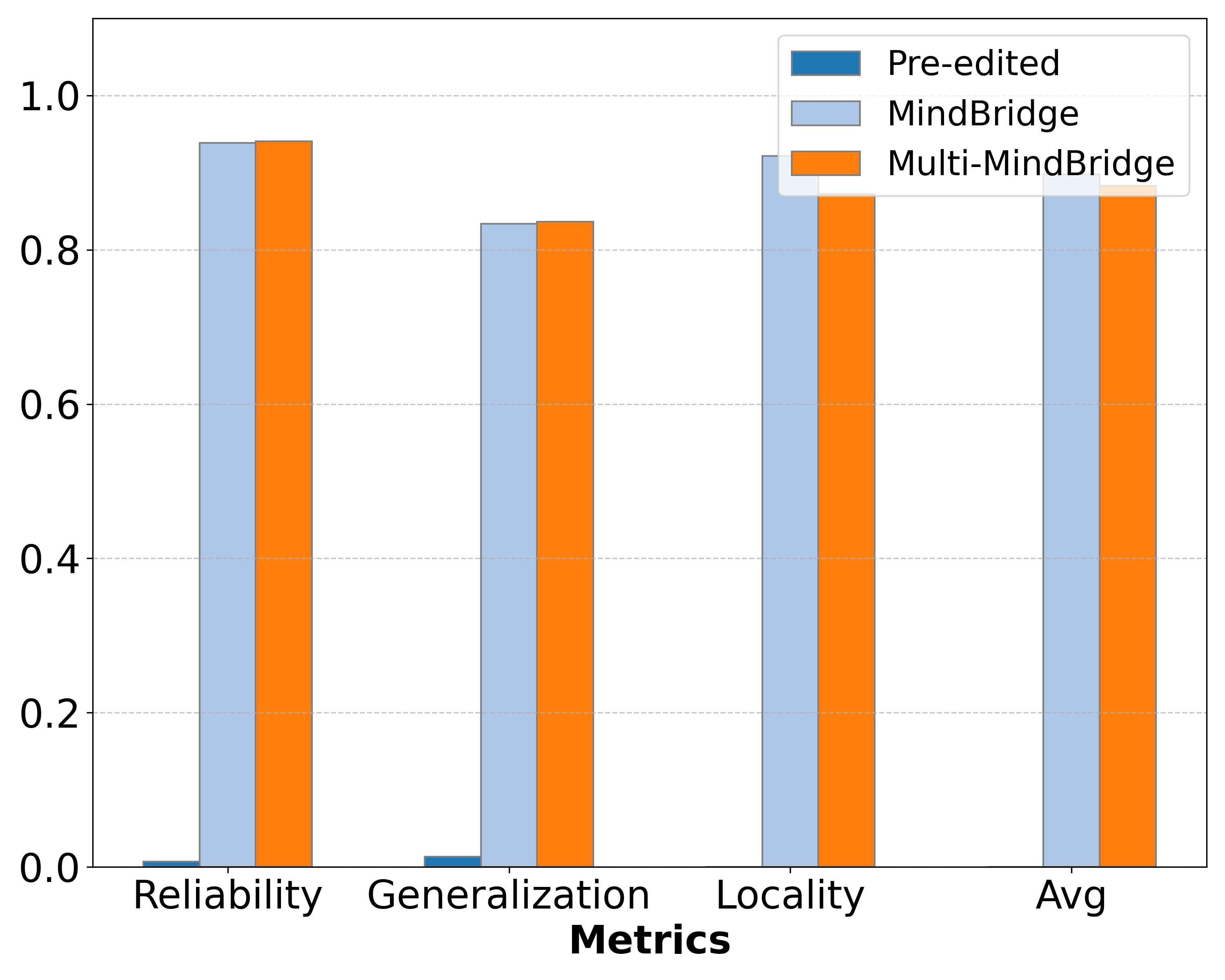}
}
\hspace{-0.01\textwidth}\subfigure[ZsRE]{ \includegraphics[width=0.22 \textwidth]{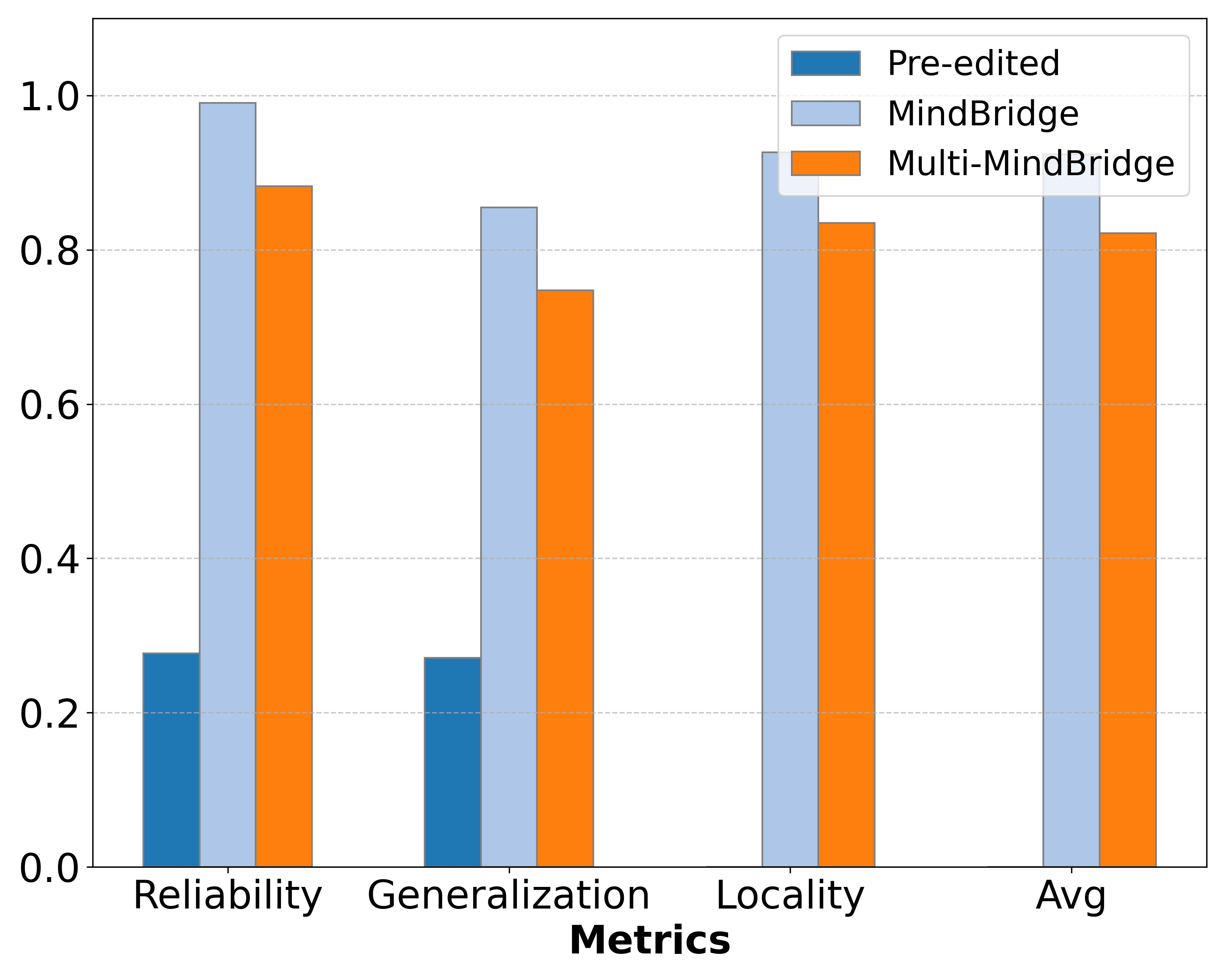}
}
\caption{Editing performance of Multi-MindBridge, which simultaneously bridges two memory modality encoders trained on different datasets and evaluates them on a single dataset.}
\label{fig:multi-mindbridge}
\end{figure}

\textbf{From MindBridge to Multi-MindBridge.} Just as multimodal large language models are not limited to one or two modalities and can simultaneously support multiple modalities \cite{shu2023audio,zhang-etal-2023-video}, we propose Multi-MindBridge, which explores the editing performance of bridging multiple distinct memory modality encoders to the same LLM. We bridge encoders pretrained on the Counterfact and ZsRE datasets (each with 10,000 edits) to the LLaMA3 (8B) model. The results are shown in Figure \ref{fig:multi-mindbridge}. It can be observed that compared to using a single encoder, the editing performance only slightly decreases but enables the LLM to simultaneously acquire the knowledge memories from both memory modality encoders. For more implementation details and exploratory experiments of Multi-MindBridge, please refer to Appendix \ref{sec:Implementation Details and Further Experiments}.

\section{Conclusion}
In this paper, we propose MindBridge, a scalable cross-model knowledge editing method designed to address the issue that most current knowledge editing methods are overfitted to a single model, leading to the problem of discarded edited knowledge and frequent re-editing with each model update. Based on the novel concept of memory modality, MindBridge enables edited knowledge to transcend individual models. Extensive  experiments conducted on two popular knowledge editing datasets and various LLMs demonstrate the effectiveness and scalability of MindBridge.

\section{Limitations}
Although MindBridge has demonstrated promising results in cross-model knowledge editing, it still faces several limitations. One limitation stems from resource constraints. We were unable to conduct tests on larger-scale models, restricting our experiments to models with up to 8B parameters. Furthermore, our focus was primarily on factual knowledge. We did not delve into other forms of knowledge, such as conceptual knowledge, which we leave this for future work.

\bibliography{custom}

\appendix

\section{Experimental Setup}\label{sec:Implementation Details}
In this section, we provide a more detailed introduction to the experimental setup, including the datasets used, a detailed explanation of the evaluation metrics, and a thorough description of the baselines.

\subsection{Datasets}

\textbf{ZsRE.} ZsRE \cite{zsre-dataset} (Zero-Shot Relation Extraction) is a question-answering dataset widely used in knowledge editing tasks. Each data entry includes a question, the subject of the question, the updated answer, rephrased questions for testing the generalization of edits, and unrelated questions for testing the locality of edits. In the experiments comparing with baselines (see Section \ref{sec:main results}), we randomly selected 10,000 samples from this dataset as $D_{edit}$. Due to the limited amount of in-scope data $\mathcal{I}$ and to prevent contamination of test data, we directly used $D_{edit}$ as a substitute for $\mathcal{I}$ during the training of MindBridge. The remaining samples that did not require editing were used as out-of-scope data $\mathcal{O}$. For other baseline methods, the edit data used were identical to those of MindBridge.

\textbf{Counterfact.} Counterfact \cite{ROME} is a more challenging knowledge editing dataset compared to ZsRE. It consists of incorrect facts that initially receive much lower scores than correct facts. Each entry in Counterfact includes a subject, an attribute of the subject to be edited, questions about attributes of non-identical subjects for testing edit locality, and paraphrases for testing edit generalization. The construction of the editing data is similar to that of ZsRE.

\subsection{Metrics}
In this paper, we use three evaluation metrics—reliability, generalization, and locality—to represent performance. For simplicity, they are abbreviated as Rel., Gen., and Loc., respectively. Their specific formulas are as follows, where $\mathbbm{1}(\cdot)$ denotes the indicator function.
\begin{align*}
    Rel. &= \frac{1}{|D_{edit}|} \sum_{(s,r,o^*) \in D_{edit}} \mathbbm{1}(F^*(s,r) = o^*), \\
    Gen. &= \frac{1}{|\mathcal{I}|} \sum_{(s,r,o^*) \in \mathcal{I}} \mathbbm{1}(F^*(s,r) = o^*), \\
    Loc. &= \frac{1}{|\mathcal{O}|} \sum_{(s,r,o^*) \in \mathcal{O}} \mathbbm{1}(F^*(s,r) = F(s,r)). \\
\end{align*}
Here, Rel. measures the proportion of edits that have been successfully applied to a model, Gen. indicates the proportion of edits to which the model generalizes after editing, and Loc. reflects the extent to which the edited model retains knowledge of unrelated facts. Note that the Loc. metric is only evaluated on the edited model. Higher values for all three metrics indicate better editing performance.

\subsection{Baselines}
Here, we introduce several knowledge editing baselines that are compared in this paper. We utilize EasyEdit\cite{easyedit} for the reproduction of these baselines. For the selection of hyperparameters among them, we follow the default settings provided in the code.
\begin{itemize}
    
        \item \textbf{FT-L} is a fine-tuning method proposed by \cite{ROME}. FT-L directly fine-tunes the feed-forward network (FFN) of a specific layer, identified by causal tracing in ROME, by maximizing the probability of all tokens in the target sequence through last token prediction.

                \item \textbf{GRACE}\cite{grace} is a lifelong editing method. It writes new mappings into a pre-trained model's latent space, creating a discrete, local codebook of edits without altering model weights. The phenomenon of GRACE's generalization collapse under extensive editing has been evidenced and discussed in \cite{wise}.
    
        \item \textbf{WISE}\cite{wise} is a method specifically designed for lifelong model editing. It inserts side memory in the FFN layers to preserve edited memory and trained a router to select which memory module to activate. To further enhance the support for continual editing, WISE incorporated a knowledge-sharding mechanism to enable different edits to be maintained in distinct parameter subspaces.

                \item \textbf{r-ROME} is an improvement upon ROME. Prior work has demonstrated that ROME can suffer from disabling edits, leading to immediate model collapse\cite{gupta2024model}. r-ROME identifies that this is caused by irregularities in ROME's implementation, specifically the asymmetric usage of key-vectors in its update equation, and proposes a more stable implementation, r-ROME.

                \item \textbf{MEMIT} \cite{memit} is a scalable multi-layer update algorithm that employs explicitly calculated parameter updates to insert new memories. Building upon the direct editing approach of ROME, it designs an edit-distribution algorithm to distribute parameter updates uniformly across multiple layers of parameters. This enables MEMIT to update thousands of new pieces of knowledge.
    
        \item \textbf{EMMET} \cite{emmet} unifies two editing methods, ROME and MEMIT, under a single optimization objective, namely, the preservation memorization objective. Furthermore, it improves upon ROME by employing equality constraints to support batched editing, achieving comparable performance to MEMIT.

                \item \textbf{AlphaEdit} \cite{alphaedit} extends the locating-and-editing method by projecting the perturbation introduced during the editing process onto the null-space of the knowledge to be preserved. Subsequently, it applies this projection to the model parameters. This mechanism ensures the model's preservation of its original knowledge following the edit.

\end{itemize}

\section{More Experimental results}

\subsection{Impact of Different Memory Modality Encoders}
We uses BERT-Base (110M) as the default modality encoder. To evaluate how different-sized BERT models affect editing performance, we compared DistillBERT (66M) \cite{distilbert} and BERT-Large (340M). As shown in Table \ref{tab:impact of different encoder}, DistillBERT delivers strong performance across GPT-J and LLaMA3. In most cases, BERT-Base and DistillBERT outperform the others, with DistillBERT achieving the best overall results on Counterfact and BERT-Base excelling on ZsRE. Despite its larger size, BERT-Large performs slightly worse than the other two models.

\begin{table*}[t]\label{Baseline table}
\centering
\small
\begin{tabular}{cc|cccc|cccc}
\toprule
\multirow{2}{*}{\textbf{Modality Encoder}} & \multirow{2}{*}{\textbf{Model}} &
\multicolumn{4}{c|}{\textbf{Counterfact}} & 
\multicolumn{4}{c}{\textbf{ZsRE}} \\
\cmidrule(lr){3-6} \cmidrule(lr){7-10} & & \textbf{Rel.$\uparrow$} & \textbf{Gen.$\uparrow$} & \textbf{Loc.$\uparrow$} & \textbf{Avg.$\uparrow$} &  
\textbf{Rel.$\uparrow$} & \textbf{Gen.$\uparrow$} & \textbf{Loc.$\uparrow$} & \textbf{Avg.$\uparrow$} \\

\midrule
DistilBERT (66M) & \multirow{3}{*}{\rotatebox{90}{GPT-J}}&   \textbf{98.21} &\textbf{ 90.64} & 92.50 & \textbf{93.79}  & 97.60 & 82.47 & 96.52 & 92.20\\
BERT-Base (110M)  &  & 94.71 & 83.57 & \textbf{93.61} & 90.63 & \textbf{99.09} & \textbf{84.22} & 95.95 & \textbf{93.09} \\
BERT-Large (340M) &  & 94.43 & 81.86 & 91.07 & 89.12 & 96.16 & 79.29 & \textbf{97.22} & 90.89 \\

\midrule

DistilBERT (66M)  & \multirow{3}{*}{\rotatebox{90}{\scriptsize {LLaMA3}}}& \textbf{97.71} & \textbf{89.31} &  90.17 & \textbf{92.39}  & 97.12  & 84.46  & 93.81  & 91.80\\
BERT-Base  (110M)  & & 93.97  & 83.58 & \textbf{92.08} & 89.88 & \textbf{98.50} & \textbf{85.02} & 92.44 & \textbf{91.99}\\
BERT-Large (340M)  &  & 92.80  & 80.81  & 89.99 & 87.87 & 96.43 & 83.15 & \textbf{95.34} & 91.64\\

\midrule
\end{tabular}
\caption{Impact of different modality encoders on editing performance, 10,000 edits. Best results are shown in bold.}
\label{tab:impact of different encoder}
\end{table*}

\subsection{Implementation Details and Further Experiments of Multi-MindBridge}\label{sec:Implementation Details and Further Experiments}
In Section \ref{sec:Further Analysis}, inspired by the idea that multimodal large language models are not restricted to one or two modalities, we proposed the implementation of Multi-MindBridge. This allows LLMs to be jointly bridged by multiple memory modality encoders while possessing the corresponding knowledge memories. Here, we elaborate on its implementation details.

Given $ n $ pre-trained memory modality encoders $ E_m^1, E_m^2, \dots, E_m^n $ with different knowledge memories and their corresponding memory modality-language modules $ P_m^1, P_m^2, \dots, P_m^n $, similar to Equation \ref{equation edited model}, the model $ F_{{multi}}^* $ after editing with Multi-MindBridge can be expressed as follows:

\begin{equation}\label{equation multi-mindbridge}
\begin{aligned}
    x_{multi} &= \bigoplus_{i=1}^{n} P_m^i(E_m^i(s, r)), \\
    F_{multi}^*(s, r) &= F(x_{multi} \oplus (s, r)).
\end{aligned}
\end{equation}

Here, $\bigoplus$ denotes the concatenation of outputs from the modality encoders $E_{m}^{1}$ to $E_{m}^{n}$ to form $x_{multi}$. The resulting $x_{multi}$ is then provided as a soft prompt to the LLM, together with the text prompt. Notably, in Multi-MindBridge, each memory modality encoder is independently trained using the same pipeline as MindBridge, and subsequently, they are combined and integrated into the edited LLM.

\begin{figure}
    \centering
    \includegraphics[width=0.8\linewidth]{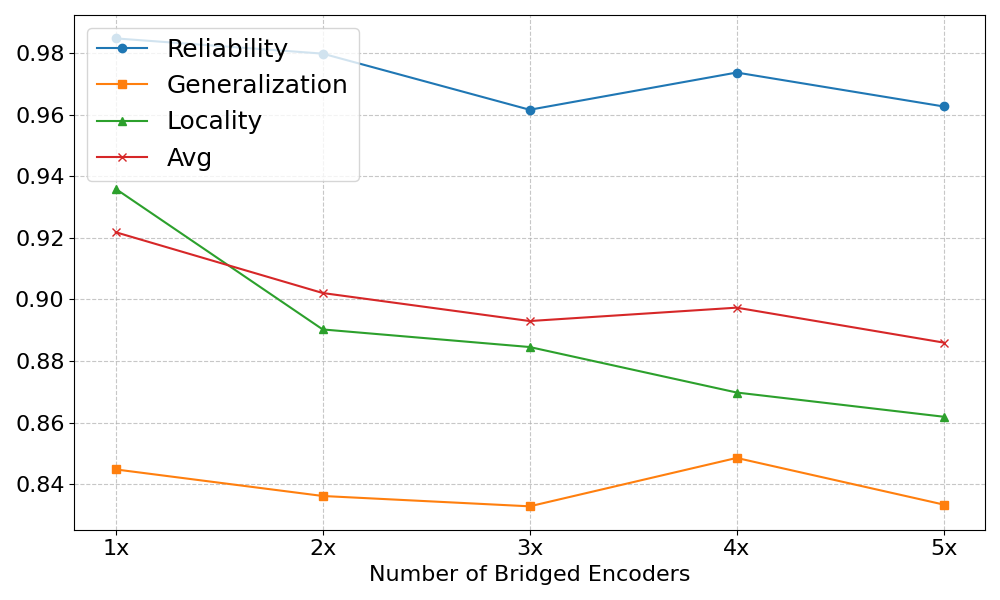}
    \caption{Editing performance of multiple memory modality encoders bridged to LLaMA3 on the ZsRE dataset, where each encoder handles 10,000 edits (non-overlapping).}
    \label{fig:multi_minbridge_with_num}
\end{figure}

In Section \ref{sec:Further Analysis}, we tested the performance of Multi-MindBridge when bridging modality encoders trained on different datasets. The results showed that compared to using a single encoder, the editing performance only slightly decreased but allowed the LLM to simultaneously acquire the knowledge memory from two encoders. Here, we test the editing performance of bridging multiple encoders trained on the same dataset. Specifically, we selected the ZsRE dataset and randomly sampled 50,000 edits, training one memory modality encoder for every 10,000 edits (with no overlap between edits). We evaluated the editing performance of bridging 1 to 5 encoders simultaneously on LLaMA3. The experimental results are shown in Figure \ref{fig:multi_minbridge_with_num}. As can be observed, as the number of simultaneously bridged encoders increases, the editing performance gradually decreases but remains satisfactory. Even when bridging up to 5 encoders, the Avg score still reaches 88.59\%.

\subsection{Impact of Editing Volume on General Capabilities}
\begin{figure}
    \centering
    \includegraphics[width=0.9\linewidth]{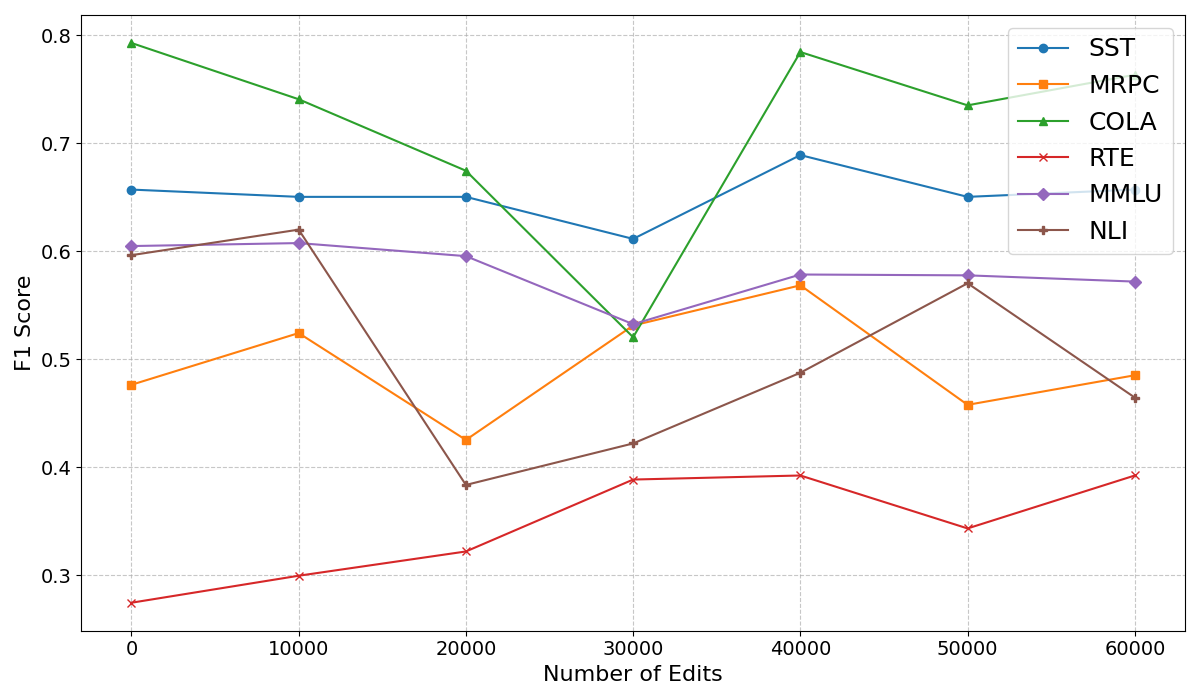}

            \caption{Change in F1-score on the GLUE benchmark for the edited LLaMA3 (8B) as the number of edited facts varies. 0 denotes the pre-edited model. Evaluation includes six tasks: SST, MRPC, CoLA, RTE, MMLU, and NLI.}
    \label{fig:glue_with_num}
\end{figure}

\cite{gupta2024model} observed that when using some model editing methods, the general capabilities of the model tend to decline as the number of edits increases. Here, we test whether MindBridge exhibits a similar phenomenon. We conduct experiments using LLaMA3 (8B), gradually increasing the number of edits from 10,000 to 60,000. Similar to Section \ref{sec:Further Analysis}, we evaluate its performance on the GLUE benchmark, which includes six tasks: SST, MRPC, CoLA, RTE, MMLU, and NLI. The experimental results are shown in Figure \ref{fig:glue_with_num}. As can be seen, with the increase in the number of edits, the F1 scores for all tasks except NLI and CoLA remain stable, and in some cases, even outperform the pre-edited model. For NLI and CoLA, the F1 scores exhibit noticeable fluctuations as the number of edits increases, but they do not show a consistent downward trend. Overall, MindBridge demonstrates good scalability in preserving the model's general capabilities.

\end{document}